\pdfoutput=1

\documentclass[11pt]{article}

\usepackage[]{acl}

\usepackage{times}
\usepackage{latexsym}
\usepackage{graphicx}
\usepackage{makecell,rotating,multirow,diagbox}
\usepackage{bbding}
\usepackage{pifont}
\usepackage{wasysym}
\usepackage{amssymb}
\usepackage[title]{appendix}
\usepackage{geometry}
\usepackage{booktabs}

\usepackage[T1]{fontenc}

\usepackage[utf8]{inputenc}

\usepackage{microtype}

\title{Qtrade AI at SemEval-2022 Task 11: An Unified Framework for Multilingual NER Task}

\author{Weichao Gan\footnotemark[1] \footnotemark[2] ,
Yuanping Lin\footnotemark[1] \footnotemark[2]  ,
Guangbo Yu\footnotemark[1] \footnotemark[2] ,
Guimin Chen \footnotemark[2] \and Qian Ye \footnotemark[4]\\
        \footnotemark[2] Qtrade AI Lab, Shenzhen China  \\
        \footnotemark[4] Guangdong Power Grid Corporation, Guangdong China  \\
         \texttt{\{waiciugam, yuenpinglynn, ygbusc\}@gmail.com} \\
         \texttt{chenguimin@foxmail.com,}\texttt{ eason\_sysuye@163.com } 
        }

\begin{document}
\maketitle
\footnotetext[1]{These authors contributed equally to this work.}

\begin{abstract}
This paper describes our system, which placed third in the Multilingual Track (subtask 11), fourth in the Code-Mixed Track (subtask 12), and seventh in the Chinese Track (subtask 9) in the SemEval 2022 Task 11: MultiCoNER Multilingual Complex Named Entity Recognition. Our system's key contributions are as follows: 1) For multilingual NER tasks, we offer an unified framework with which one can easily execute single-language or multilingual NER tasks, 2) for low-resource code-mixed NER task, one can easily enhance his or her dataset through implementing several simple data augmentation methods  and 3) for Chinese tasks, we propose a model that can capture Chinese lexical semantic, lexical border, and lexical graph structural information. Finally, our system 
achieves macro-f1 scores of 77.66, 84.35, and 74.00 on subtasks 11, 12, and 9, respectively, during the testing phase.
\end{abstract}

\section{Introduction}
SemEval 2022 Task 11: MultiCoNER Multilingual Complex Named Entity Recognition\cite{multiconer-report} focuses on extracting semantically ambiguous complex named entities\cite{meng2021gemnet}  in short, low-context and code-mixed\cite{fetahu2021gazetteer} scenarios. The domain adaptability capacity of the system is in high demand for this shared task, which contains 13 tracks in English, Spanish, Dutch, Russian, Turkish, Korean, Farsi, German, Chinese, Hindi, Bangla, multi-language, and code-mixed.  Among them, multi-language, and code-mixed tracks are in the mixed language, while the other tracks are in the single language.  The difference between multi-language track and code-mixed track is that the corpus of multi-language track is multilingual, but the words in each sentence are in a single language, while the corpus of code-mixed track is a corpus in which the words or phrases in each sentence may be from different languages, for example the sentence I Liebe Sie (I love you) consists of English, French and German. 
Furthermore, the datasets offered by the organizers , which may mainly from Wikipedia, web questions and user queries, comprise data from 11 languages, with each language containing around 15,000 training samples and 800 development (dev) samples\cite{multiconer-data}. The corpus contains a total of six entity categories, namely location (LOC), person (PER), product (PROD), group (GRP), corporation (CORP) and creative works (CW).  
The main contributions of our system are as follows: 
1) we propose a unified framework for multilingual NER tasks, using which one can easily perform monolingual or multilingual NER tasks;
2) for low-resource code-mixed NER tasks, we provide several simple and effective data augmentation methods to easily increase the amount of data; 
3) for Chinese tasks, we propose a model that captures Chinese lexical semantics, lexical boundaries and lexical graph structure information in a model.

\section{Related work}
Named Entity Recognition (NER)—a classic and fundamental task that aims to extract named entities from a sentence—plays an important role in a variety of downstream tasks in the field of NLP, including relation extraction \cite{zhong2021frustratingly}, knowledge graph construction \cite{bosselut2019comet}, question answering \cite{diefenbach2018core} and so on. 
For a long time, the development of NER was slow, especially before the rise of neural networks, and NER mostly used statistical machine learning methods like HMM\cite{morwal2012named} and CRF\cite{konkol2013crf}.  Although these methods were effective at the time, they were still stretched for complex scenarios. 
After the rise of neural networks, especially structured networks such as RNNs\cite{sherstinsky2020fundamentals} and CNNs\cite{lecun1998gradient}, NER has been greatly developed, and the accuracy has been greatly improved compared with traditional statistical learning methods.  There are also methods that combine neural networks with traditional statistical learning methods, such as BiLSTM+CRF\cite{chen2017improving}.
Although these methods take NER to another higher level, they also have certain problems. For example, RNNs cannot model bi-directionally context, and then there is the inability of RNNs to capture long-term dependencies, and though CNNs can model bi-directionally, but due to the size of convolutional kernels, they can only model local contextual information at the same time and cannot capture global contextual information.
Recently, self-training pre-training models on large-scale corpus such as BERT\cite{kentonbert} and its variant versions have greatly improved the accuracy of NER tasks and effectively solved the problem that RNNs cannot capture long-range dependencies as well as dynamic contexts. 
From statistical models like HMM and CRF to deep learning models like CNN, RNN, and transformer-based pre-trained models, the accuracy of NER tasks is increasing and has reached commercial levels in many scenarios\cite{yadav2018survey}. 
Despite its remarkable progress, NER still faces some challenges\cite{ma2020simplify}, such as the problem of discontinued and nested named entities\cite{yan2021unified}, the challenge of cross-domain and cross-lingual transfer learning\cite{mueller2020sources}, the lack of lexicon information when using sub-character tokenization strategy and word boundary information of Chinese NER tasks\cite{zhang2018chinese}, and the ambiguity of named entities\cite{meng2021gemnet}  under different semantic circumstance.

\section{System overview}
\begin{figure}[ht]
\centering
\includegraphics[scale=0.35]{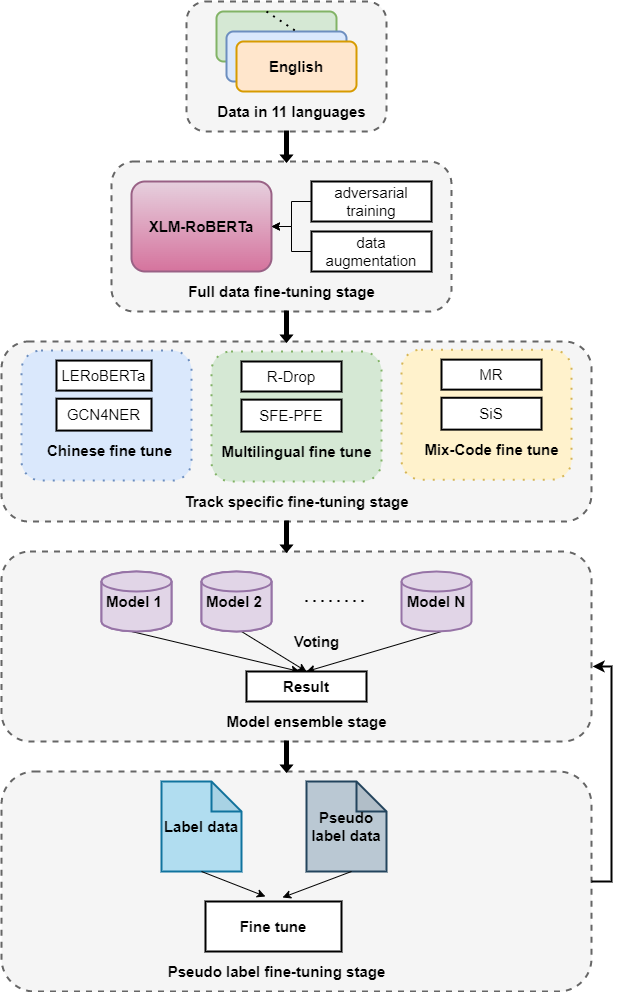}
\caption{The architecture and procedures of our system}
\label{fig2-1}
\end{figure}

Figure \ref{fig2-1} depicts our system's overall architecture and technological process. As we can see, the unified framework allows us to complete all of the subtasks.

\subsection{Backbone encoder}
We selected XLM-RoBERT-large \cite{conneau2020unsupervised} as our backbone encoder to establish framework unity and make full use of data from different tracks to realize language transfer. XLM-RoBERTa is a multilingual version of RoBERTa that has been pre-trained on 2.5TB of data in 100 languages. The large version has 24 transformer layers, 16 self-attention heads per layer, and a hidden size of 1024.

\subsection{Training procedures}
We take multiple steps to fine tune our system, including full data fine-tuning stage, track specific fine-tuning stage, model ensemble stage and pseudo label fine-tuning stage.

\subsubsection{Full data fine-tuning stage}
As the first step of our system, we fine tune our model with data from all languages provided by official. At this stage, we hope that the model can learn the distribution of datasets and capture the named entity information from the total data, which will aid our model in transferring to those data on specific tracks. To improve the model's cognitive ability and stability, we implemented the following set of training skills.
\\
\textbf{Data Augmentation:}We did not simply feed the data into our model, but enhanced data firstly in a simple way—concatenating sentences randomly, including \textbf{bisent-uni}—concatenating two sentences from the same language randomly into a new sentence, \textbf{bisent-mix}—concatenating two sentences from different language randomly into a new sentence, \textbf{mulsent-uni}—concatenating several sentences from the same language randomly into a new sentence  with a length of no more than 512 and \textbf{mulsent-mix}—concatenating several sentences from different language randomly into a new sentence  with a length of no more than 512.
\\
\textbf{Adversarial Training:} Considering that the deep neural networks are vulnerable to adversarial examples, we adopt FGM\cite{miyato2017adversarial}, PGD\cite{mkadry2017towards} and FreeLB\cite{zhu2019freelb} adversarial training techniques to keep our model tolerant of adversarial examples and more robust.  These methods create adversarial examples by adding a small perturbation to input that maximizes the loss. The formulation of adversarial training can be drew below,
\begin{equation}
    \min_{\theta}\mathbb{E}_{(x,y)\sim D}\left[ \max_{\delta\in S} L(\theta, x+\delta,y) \right]
\end{equation}
where $\delta$ is a small perturbation.  The inner maximization can be solved by projected gradient ascent and the outer minimization can be solved by gradient descent. PGD and FreeLB are the improved version of FGM, both of which  are devoted to find a more reasonable perturbation. 

\subsubsection{Track specific fine-tuning stage}
\begin{figure}[ht]
\centering
\includegraphics[scale=0.4]{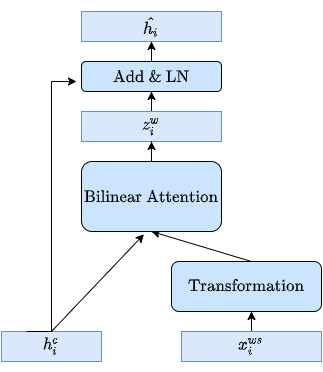}
\caption{Structure of Lexicon Adapter. This structure pays bilinear attention to characters and vocabulary at the same time, weights vocabulary features into vectors, and then adds them to the input character level vector, and then performs layer normalization.}
\label{fig2-2}
\end{figure}

At this stage, we use different methods to fine tune according to the  specific track.
\\
\textbf{Track 9 - Chinese (ZH):}
Track 9 is devoted to the recognition of Chinese named entities. There are significant differences between Chinese and Indo-European languages like English, German, and Spanish. For example, there is no word boundary in the Chinese lexicon, however Chinese word semantic and boundary information is useful for NER. Therefore, after the fine-tuning at the first stage, we utilize some techniques to improve our model's ability to understand Chinese. In particular, we incorporate LEBERT\cite{liu2021lexicon} and CoGraph4NER\cite{sui2019leverage} into our framework to improve our model's understanding of Chinese lexicon boundaries and semantics. It is worth noting that, we did not directly implement these methods; instead, we drew lessons from the ideas presented in these two papers and appropriately transformed the structures mentioned in these two papers to apply to our system.
\\
\textbf{LERoBERTa:}
LERoBERTa is our modified version of LEBERT that integrates lexicon information into the specific layers of the pre-training model to obtain word-related information via a structure called Lexicon Adapter. The specific structure of Lexicon Adapter is shown in Figure \ref{fig2-2},

\begin{figure}[ht]
\centering
\includegraphics[scale=0.4]{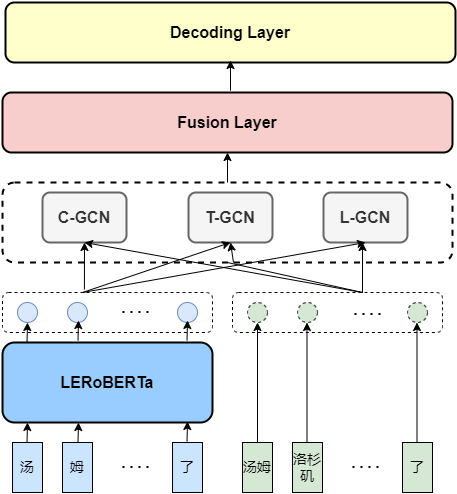}
\caption{Main architecture of LERoBERTa-GCN4NER. Characters of "Tom left Los Angeles Airport" are encoded by LERoberta and then the hidden vectors and vectors of lexicons are inputted into the GCN module, the outputs of which will then be fused by weighted summation in fusion layer. Finally, the decoding layer assigns a label to each character.}
\label{fig2-3}
\end{figure}

Because the original LEBERT paper is applied to BERT, it is not consistent with our framework. In order to apply it to our framework, we converted it to LERoBERTa, which differs slightly from LEBERT.
\\
\textbf{LERoBERTa-GCN4NER:}
LERoBERTa does not fully utilize the graph relation of containing, transition, and lattice between words and characters because it only integrates information from the bottom encoding layer. An assumption is that the characters in the sentence can capture the boundaries and semantic information of self-matched lexical words using the containing graph (C-graph), the transition graph (T-graph) can assist the character in capturing the semantic information of the nearest contextual lexical words implicitly, and the lattice graph (L-graph) can capture some information of self-matched lexical words explicitly. So, we make some improvements to CoGraph4NER, called GCN4NER, to make full use of the graph relation and thus improve the model's ability to capture Chinese lexicons. GCN4NER is a modified version of CoGraph4NER that replaces the LSTM encoder with LERoBERTa to conform to our algorithm framework and replaces the GAT module with GCN to improve calculation speed. The overall system of LERoBERTa-GCN4NER is shown in Figure \ref{fig2-3}, 
\\
\textbf{Track 12 - Multilingual:}
Track 12 focuses on multilingual named entity recognition and the difficulty with this track is that it contains multiple languages, each with its own syntax. So the model must learn multiple language characteristics concurrently. It is difficult for participants to analyze the model's bad cases because they are not  familiar to every language. At this stage,
we not only use R-Drop\cite{wu2021r} but also utilize shared feature extractors and private feature extractors\cite{chen2019multi}, considering the fact that there are differences in grammar and common characteristics in semantics of texts in different languages.
\\
\textbf{R-Drop:}
Since deep neural networks are very prone to overfitting, the Dropout method randomly discards some neurons in each layer to avoid overfitting during training. 
Based on this idea, researchers improve dropout method, called R-Drop.
Given a training sample $D=\{x_i, y_i\}, i=1,\cdots,n$, for each sample , it goes through the forward feedback of two different sub-networks to get two predicted probabilities $P_1$ and $P_2$. Although the two sub-networks come from the same model, they are not exactly the same because Dropout randomly discards some neurons, and  $P_1$ is not equal to $P_2$.
\\
\textbf{Shared Feature Extractor and Private Feature Extractor:}
We leveraged the ideas of the paper\cite{chen2019multi} and modified the model in combination with datasets offered by the official. With this method, our model can learn general and specific characteristics between different language which may improve model's comprehension.

The shared feature extractor consists of a shared feature learner and a language discriminator. The shared feature learner extracts general features from different languages, and then inputs them into the language discriminator to judge which language these features belongs to. When the language discriminator is unable to determine which language the current text belongs to, it signifies that the shared feature learner has learnt the commonalities between the languages, hence fulfilling the goal of perplexing the language discriminator.
\\
There are differences in grammar and meaning of texts in different languages, so the model of this scheme also designs a private feature extractor. The private feature extractor performs feature extraction on the word vector output by backbone encoder, and then outputs it to the multilayer perceptron (MLP) of each language. The MLP of each language extracts the exclusive language features in the text, and then splices the output of each language.
\\
\textbf{Track 13 - Code-Mixed:}
Track 13 is a track dedicated to code-mixed data. The prevalence of code-mixed text is fast increasing on social media platforms such as Twitter. The code-mixed task is difficult because it introduces a large number of unseen constructions as a result of merging the lexicon and syntax of two or more languages, and the available data is insufficient in comparison to the other sub-tasks. Therefore, after fine-tuning at the first state, we utilized data augmentation\cite{dai2020analysis} to supplement the training data in this task. When the labeled datasets are insufficient, data augmentation is a frequently used strategy for enhancing generalization. However, the NER task is concerned with sequence labeling at the token level, and the majority of data augmentation methods at the sentence level may compromise label integrity. We used \textbf{Mention replacement (MR)} and \textbf{Shuffle within segments (SiS)} techniques to resolve the issue. MR is a data  augmentation approach that replaces the position of entities in the original sentence with other entities of the same category while keeping the non-entity part of the sentence unchanged. And SiS is a data enhancement method that keeps the order of words of entities in the original sentence unchanged and disrupts the order of words in the non-entity part.

\subsection{Model ensemble stage}
Model Ensemble is a method for integrating multiple trained models in order to improve the system's generalization ability in test datasets\cite{allen2020towards}. Model ensemble can be achieved in a variety of ways, including voting, averaging, stacking, and confidence blending. Voting is one of them, and it is a simple yet effective method, so we use it in our system. We trained several models for each track based on different hyper parameters and then predicted test datasets to get corresponding results. We then voted for the final result based on these intermediate results. 

\subsection{Pseudo label fine-tuning stage}
Pseudo labelling is a type of semi-supervised learning in which the model trained by labelled data is used to generate pseudo labels for an unlabelled dataset\cite{lee2013pseudo}. We then put both the original labelled dataset and a portion of the unlabelled dataset with pseudo labels into our models for final training during the pseudo label fine-tuning stage. Voting is how we generate pseudo labels. We choose sentences that can be consistently predicted by all models generated in the model ensemble stage as training samples with reasonable pseudo-labels.

\section{Experimental setup}
In this session, we are going to describe the implementation details of our system.
\\
\textbf{Dataset and word embeddings:} Train and dev datasets we used are provided by official and we don’t use test dataset before testing phase because it was not available.  Besides, experiments at model ensemble stage and pseudo label fine-tuning stage are carried on test dataset.Since LERoBERTa and GCN4NER techniques require word embedding and considering the lack of embedding of relevant words in the pre-trained model XLM-RoBERTa, we use word embedding sgns.merge.word \footnotemark[2]
\footnotetext[2]{https://drive.google.com/file/d/1Zh9ZCEu8\_eSQ-qkYVQufQDNKPC4mtEKR/view.} trained by skip-gram. 
\\
\textbf{Processing and hyper parameters:}
Limited by the length of the paper, we will only  briefly introduce our main processing and part of hyper parameters, with which our system can reach the best result. Learning rate was set to $1 \times 10^{-6}$, warming up proportion to 0.06, drop out rate to 0.2,  batch size to 32, epoch to 30, random seed to 42 and the number of voting models to 7. Augmentation method for full data fine-tuning stage was \textbf{mulsent-uni}, decoder strategy was softmax and adversarial training method was PGD with $\epsilon=1.0,\alpha=0.1,K=3$. For more details, please check the Appendix \ref{appendix1}. Through out our system we almost only use softmax as our decoder strategy because we have found CRF has no effect on improving the scores compared to softmax but consumes more computing resources and storage space. So, we drop CRF strategy. One plausible explanation for this is that pre-trained models have already caught the relations between tokens which may not be well captured by non-pretrained models. We selected the best epoch and the best hyper parameters using performance (measured in terms of macro-f1 score) on corresponding dev dataset.

\section{Results}
In this section, we will report our main experiment and provide an analysis of the results. Unless otherwise specified, the hyper parameters in our experiments are configured in accordance with Section 3.
\\
\textbf{Full data fine-tuning stage:}
Table \ref{tab.fd} displays the results of experiments performed with the following parameters: FGM with $\epsilon =0.8$, PGD with $\epsilon=1, \alpha=0.1, K=3$ and FreeLB with $adv\_lr=0.3, mag=0.05, K=3$. These hyper parameters are the best that we have found. 

\begin{table*}[h]
\caption{Marco-f1 scores(\%) on  dev dataset under different  methods at full data fine-tuning stage }
\centering
\renewcommand\arraystretch{1.2}{
\setlength{\tabcolsep}{4.5mm}{
\begin{tabular}{c c| c c c}
\toprule[1pt]
\multicolumn{2}{c|}{methods} 
& Chinese(ZH)  & Multilingual & Mix-Code \\ 
\hline
$\backslash$ & $\backslash$ & 86.31       & 85.82        & 75.93 \\
\hline
\multirow{4}{*}{data augmentation} 
 & bisent-uni & 86.20       & 86.02        & 77.01 \\
 & mulsent-uni & 86.06       & 86.06        & 77.19 \\
 & bisent-mix & 85.27       & 86.01        & 74.67 \\
 & mulsent-mix & 86.50       & 85.91        & 74.43 \\
\hline
\multirow{3}{*}{adversarial training} 
 & FGM              & 86.85       & 86.85        & 75.54 \\
 & PGD              & 87.37       & 86.33        & 75.96 \\
 & FreeLB           & 86.81       & 86.14        & 75.88 \\
\hline
methods ensemble  & mulsent-uni+PGD  & \textbf{88.02}       & \textbf{86.93}        & \textbf{78.21} \\
\bottomrule[1pt]
\end{tabular}}}
\label{tab.fd}
\end{table*}

As we can see from Table \ref{tab.fd}, three tracks have an improvement in the mulsent-uni data augmentation method where Chinese(ZH) and  Multilingual tracks improve nearly 0.5 percent and Mix-Code tack increases 1.2 percent to 77.19. What’s more, all adversarial training techniques we employed in our system make a great improvement in Chinese(ZH) and Multilingual track, but nearly have no effect on Mix-Code track.  However, when we adopt mulsent-uni and PGD simultaneously, scores are surprisingly high, 88.46 for ZH-Chinese, 86.93 for Multilingual and 78.21 for Mix-Code and this checkpoint is our best checkpoint at  full data fine-tuning stage which will continue to be used at track specific fine-tuning stage. 

\textbf{Chinese (ZH) track fine-tuning:}
As for Chinese (ZH) track, we employ LERoBERTa, GCN4NER and LERoBERTa-GCN4NER for fine-tuning.
Table \ref{tab.zhfinetune} demonstrates that, LEBERT gets the lowest score without our full data fine-tuning stage and our modified version, LERoBERTa, GCN4NER and LERoBERTa-GCN4NER have a positive effect on improving macro-f1 score. We can prove our hypothesis that LERoBERTa only integrates the information of words from the bottom encoding layer, and that it does not make full use of the graph relation of containing, transition, and lattice between words and characters, by observing the differences in results between our modified models.  As a result, LERoBERTa just climbs by 0.18 percent. Similar to GCN4NER, it may just consider character graph relationships while ignoring lexical semantic information. So, GCN4NER increases by 0.32 percent. When these two approaches are combined, we discover that LERoBERTa-GCN4NER improves by 1\%, giving it the highest score on the dev dataset in our system.

\begin{table*}[h]
\caption{Marco-f1 scores(\%) on dev dataset under different models at Chinese (ZH) track fine-tuning stage }
\centering
\renewcommand\arraystretch{1.5}{
\setlength{\tabcolsep}{1.8mm}{
\begin{tabular}{c c c c c c}
\toprule[1pt]
Models & XLM-RoBERTa  & LEBERT & LERoBERTa & GCN4NER & LERoBERTa-GCN4NER \\ 
\hline
Chinese(ZH) & 88.46  & 86.02  &  88.62        & 88.78  &  \textbf{89.4} \\
\bottomrule[1pt]
\end{tabular}}}
\label{tab.zhfinetune}
\end{table*}

\begin{table*}[h!]
\caption{ Marco-f1 scores(\%) on on  dev dataset under different methods at multilingual track fine-tuning stage }
\centering
\renewcommand\arraystretch{1.5}{
\setlength{\tabcolsep}{7.5mm}{
\begin{tabular}{c c c c c}
\toprule[1pt]
methods & $\backslash$  & R-Drop & PFE-SFE & R-Drop+PFE-SFE  \\ 
\hline
Multilingual & 86.93  & 87.15  &  87.21        &    \textbf{87.42}    \\
\bottomrule[1pt]
\end{tabular}}}
\label{tab.mlfinetune}
\end{table*}

\textbf{ Multilingual track fine-tuning: }
This experiment continually use the best checkpoint from full data fine-tuning stage . We try R-Drop technique and  private feature extractor and shared feature extractor (PFE-SFE) method to improve the performance of our model on this track. In the R-Drop experiment, we use the average method for the KL divergence, and set coefficient parameter $\alpha$ to 0.1 and it obtains a marco-f1 score of 87.15 on the dev dataset which means that  this method has a good effect on multilingual tasks. What‘s more, we adopt PFE-SFE method and the  marco-f1 score achieve 87.21 in our experiment. 


It can be seen from Table \ref{tab.mlfinetune} that both R-Drop and PFE-SFE have improved macro f1 score compared to the baseline model, R-Drop is 0.22\% higher than the baseline, and PFE-SFE is 0.28\% higher than the baseline. It can be seen that both methods have certain effects. When we combine R-Drop with PFE-SFE, the macro-f1 value reaches 87.42, which is 0.49\% higher than the baseline.

\textbf{Mix-Code track fine-tuning:}
As for Mix-Code track, we used mention replacement (MR) and shuffle within segments (SiS) techniques to resolve the insufficiency issue of mix-code data and increase the generalization ability of our model on this track. We not only utilize the parameter setting at full data fine-tuning stage, but also employ the best checkpoint at full data fine-tuning stage.

\begin{table}[h]
\caption{ Marco-F1 scores(\%) on dev dataset under different data augmentation strategies at code-mixed fine-tuning stage }
\centering
\renewcommand\arraystretch{1.5}{
\setlength{\tabcolsep}{3.7mm}{
\begin{tabular}{c c }
\toprule[1pt]
Data Augmentation Methods & Mix-Code   \\ 
\hline
$\backslash$ & 79.8   \\
Mention replacement (MR) & 82.0   \\
Shuffle within segments (SiS) & 81.1   \\
MR + SiS & \textbf{82.2}   \\
\bottomrule[1pt]
\end{tabular}}}
\label{tab.mixfinetune}
\end{table}

As shown in Table \ref{tab.mixfinetune}, all of the data augmentation techniques we utilized increase macro-f1 over the baseline value when no augmentation is applied. The MR technique enables the model to get a more accurate representation of entity knowledge and entity boundary information based on external knowledge. Additionally, by changing the order of the sequences, the SiS technique enables the model to learn a more robust position embedding.And the result in bold is our best score on the leaderboard before testing phase.

\textbf{Model ensemble stage:}
At this stage, test data is available and we choose the top-7 models based on marco-f1 score under 7 sets of different parameters to vote on the final results. For details, please check the Appendix \ref{appendix2}. The second row in Table \ref{tab.modelensemble} shows the best results predicted singly by our chosen models. As we can see from Table \ref{tab.modelensemble}, model ensemble has a positive effect. Before the number of models increases to 7, scores increase as the number of models increases but the growth rate slows down which means more models will have little effect even negative effect. 

\begin{table}[h!]
\caption{Macro-F1 scores(\%) on test dataset under different number of models at model ensemble stage(N denotes the number of models) }
\centering
\renewcommand\arraystretch{1.5}{
\setlength{\tabcolsep}{2.4mm}{
\begin{tabular}{c c c c}
\toprule[1pt]
N & Chinese(ZH)  & Multilingual & Mix-Code  \\ 
\hline
1 &     67.71  &    73.18  &    80.32       \\
3 &     69.42  &    73.86  &    82.43       \\
5 &     \textbf{69.73}  &    \textbf{74.32}  &    \textbf{82.75}       \\
7 &    68.20  &    73.67  &    81.53       \\
\bottomrule[1pt]
\end{tabular}}}
\label{tab.modelensemble}
\end{table}

\textbf{Pseudo labeling  fine-tuning stage:}
After model ensemble stage, we select 7 models with the highest f1 values on the development set to make predictions on the test set, and choose the results that all seven models predict consistently as the pseudo-labeled data and then we put them together with train dataset for fine-tuning at this stage. 
We continually use the best checkpoint of LERoBERTa-GCN4NER to fine tune on the Chinese(ZH) track and same as the Multilingual track and Code-mixed track. In addition, the hyperparameters remain the same as in the previous phase. The results we obtained are as shown in Table \ref{tab.pseudo}. Obvious as the effect is, macro-f1 scores increases 5.22 points on the Chinese(ZH) track, 4.04 points on the Multilingual track and 3.59 points on the 
Code-mixed track.

\begin{table}[h!]
\caption{Macro-F1 scores(\%) on test dataset  at pseudo labeling fine-tuning stage }
\centering
\renewcommand\arraystretch{1.5}{
\setlength{\tabcolsep}{1.3mm}{
\begin{tabular}{c c c c}
\toprule[1pt]
pseudo & Chinese(ZH)  & Multilingual & Mix-Code  \\ 
\hline
\XSolidBrush  &     67.71  &    72.87  &    80.32       \\
\Checkmark  &   \textbf{72.93}  &    \textbf{76.91}  &    \textbf{83.91}       \\
\bottomrule[1pt]
\end{tabular}}}
\label{tab.pseudo}
\end{table}

Next, same as model ensemble stage we choose 7 sets of different parameters to train on the pseudo label dataset and get 7 sets of checkpoints. For more details about these models, please check the Appendix \ref{appendix1}. The results are shown in Table \ref{tab.pseudomodelensemble}. As we can see, the results are similar to those at model ensemble stage.  Macro-f1 scores also have an improvement but the increments are less than before.The value in bold is our result in the final ranking.

\begin{table}[htp]
\caption{Macro-F1 scores(\%) on test dataset under different number of models at pseudo labeling fine-tuning stage (N denotes the number of models) }
\centering
\renewcommand\arraystretch{1.5}{
\setlength{\tabcolsep}{2.4mm}{
\begin{tabular}{c c c c}
\toprule[1pt]
N & Chinese(ZH)  & Multilingual & Mix-Code  \\ 
\hline
1 &     72.93  &    76.91  &    83.91       \\
3 &     73.60  &    77.37  &    84.29       \\
5 &     \textbf{74.00}  &    \textbf{77.66}  &    \textbf{84.35}       \\
7 &    73.25  &    77.32  &    84.02       \\
\bottomrule[1pt]
\end{tabular}}}
\label{tab.pseudomodelensemble}
\end{table}

\section{Conclusion}
In this paper, we have introduce our system step by step which can be regarded as an universal framework for multilingual NER task. Besides, we proposed a graph-based model to make up for the lack of understanding capacity of Chinese lexicon. Adequate ablation experiments shows that our methods work for this task.  In future efforts, we plan to further improve our system to better
handle polysemous scenarios.

\bibliography{anthology,custom}

\begin{thebibliography}{29}
\expandafter\ifx\csname natexlab\endcsname\relax\def\natexlab#1{#1}\fi

\bibitem[{Allen-Zhu and Li(2020)}]{allen2020towards}
Zeyuan Allen-Zhu and Yuanzhi Li. 2020.
\newblock Towards understanding ensemble, knowledge distillation and
  self-distillation in deep learning.
\newblock \emph{arXiv preprint arXiv:2012.09816}.

\bibitem[{Bosselut et~al.(2019)Bosselut, Rashkin, Sap, Malaviya, Celikyilmaz,
  and Choi}]{bosselut2019comet}
Antoine Bosselut, Hannah Rashkin, Maarten Sap, Chaitanya Malaviya, Asli
  Celikyilmaz, and Yejin Choi. 2019.
\newblock Comet: Commonsense transformers for automatic knowledge graph
  construction.
\newblock In \emph{Proceedings of the 57th Annual Meeting of the Association
  for Computational Linguistics}, pages 4762--4779.

\bibitem[{Chen et~al.(2017)Chen, Xu, He, and Wang}]{chen2017improving}
Tao Chen, Ruifeng Xu, Yulan He, and Xuan Wang. 2017.
\newblock Improving sentiment analysis via sentence type classification using
  bilstm-crf and cnn.
\newblock \emph{Expert Systems with Applications}, 72:221--230.

\bibitem[{Chen et~al.(2019)Chen, Hassan, Hassan, Wang, and
  Cardie}]{chen2019multi}
Xilun Chen, Ahmed Hassan, Hany Hassan, Wei Wang, and Claire Cardie. 2019.
\newblock Multi-source cross-lingual model transfer: Learning what to share.
\newblock In \emph{Proceedings of the 57th Annual Meeting of the Association
  for Computational Linguistics}, pages 3098--3112.

\bibitem[{Conneau et~al.(2020)Conneau, Khandelwal, Goyal, Chaudhary, Wenzek,
  Guzm{\'a}n, Grave, Ott, Zettlemoyer, and Stoyanov}]{conneau2020unsupervised}
Alexis Conneau, Kartikay Khandelwal, Naman Goyal, Vishrav Chaudhary, Guillaume
  Wenzek, Francisco Guzm{\'a}n, {\'E}douard Grave, Myle Ott, Luke Zettlemoyer,
  and Veselin Stoyanov. 2020.
\newblock Unsupervised cross-lingual representation learning at scale.
\newblock In \emph{Proceedings of the 58th Annual Meeting of the Association
  for Computational Linguistics}, pages 8440--8451.

\bibitem[{Dai and Adel(2020)}]{dai2020analysis}
Xiang Dai and Heike Adel. 2020.
\newblock An analysis of simple data augmentation for named entity recognition.
\newblock In \emph{Proceedings of the 28th International Conference on
  Computational Linguistics}, pages 3861--3867.

\bibitem[{Diefenbach et~al.(2018)Diefenbach, Lopez, Singh, and
  Maret}]{diefenbach2018core}
Dennis Diefenbach, Vanessa Lopez, Kamal Singh, and Pierre Maret. 2018.
\newblock Core techniques of question answering systems over knowledge bases: a
  survey.
\newblock \emph{Knowledge and Information systems}, 55(3):529--569.

\bibitem[{Fetahu et~al.(2021)Fetahu, Fang, Rokhlenko, and
  Malmasi}]{fetahu2021gazetteer}
Besnik Fetahu, Anjie Fang, Oleg Rokhlenko, and Shervin Malmasi. 2021.
\newblock {Gazetteer Enhanced Named Entity Recognition for Code-Mixed Web
  Queries}.
\newblock In \emph{Proceedings of the 44th International ACM SIGIR Conference
  on Research and Development in Information Retrieval}, pages 1677--1681.

\bibitem[{Kenton and Toutanova()}]{kentonbert}
Jacob Devlin Ming-Wei~Chang Kenton and Lee~Kristina Toutanova.
\newblock Bert: Pre-training of deep bidirectional transformers for language
  understanding.
\newblock \emph{Universal Language Model Fine-tuning for Text Classification},
  page 278.

\bibitem[{Konkol and Konop{\'\i}k(2013)}]{konkol2013crf}
Michal Konkol and Miloslav Konop{\'\i}k. 2013.
\newblock Crf-based czech named entity recognizer and consolidation of czech
  ner research.
\newblock In \emph{International conference on text, speech and dialogue},
  pages 153--160. Springer.

\bibitem[{LeCun et~al.(1998)LeCun, Bottou, Bengio, and
  Haffner}]{lecun1998gradient}
Yann LeCun, L{\'e}on Bottou, Yoshua Bengio, and Patrick Haffner. 1998.
\newblock Gradient-based learning applied to document recognition.
\newblock \emph{Proceedings of the IEEE}, 86(11):2278--2324.

\bibitem[{Lee et~al.(2013)}]{lee2013pseudo}
Dong-Hyun Lee et~al. 2013.
\newblock Pseudo-label: The simple and efficient semi-supervised learning
  method for deep neural networks.
\newblock In \emph{Workshop on challenges in representation learning, ICML},
  volume~3, page 896.

\bibitem[{Liu et~al.(2021)Liu, Fu, Zhang, and Xiao}]{liu2021lexicon}
Wei Liu, Xiyan Fu, Yue Zhang, and Wenming Xiao. 2021.
\newblock Lexicon enhanced chinese sequence labeling using bert adapter.
\newblock In \emph{Proceedings of the 59th Annual Meeting of the Association
  for Computational Linguistics and the 11th International Joint Conference on
  Natural Language Processing (Volume 1: Long Papers)}, pages 5847--5858.

\bibitem[{Ma et~al.(2020)Ma, Peng, Zhang, Wei, and Huang}]{ma2020simplify}
Ruotian Ma, Minlong Peng, Qi~Zhang, Zhongyu Wei, and Xuan-Jing Huang. 2020.
\newblock Simplify the usage of lexicon in chinese ner.
\newblock In \emph{Proceedings of the 58th Annual Meeting of the Association
  for Computational Linguistics}, pages 5951--5960.

\bibitem[{M{\k{a}}dry et~al.(2017)M{\k{a}}dry, Makelov, Schmidt, Tsipras, and
  Vladu}]{mkadry2017towards}
Aleksander M{\k{a}}dry, Aleksandar Makelov, Ludwig Schmidt, Dimitris Tsipras,
  and Adrian Vladu. 2017.
\newblock Towards deep learning models resistant to adversarial attacks.
\newblock \emph{stat}, 1050:9.

\bibitem[{Malmasi et~al.(2022{\natexlab{a}})Malmasi, Fang, Fetahu, Kar, and
  Rokhlenko}]{multiconer-data}
Shervin Malmasi, Anjie Fang, Besnik Fetahu, Sudipta Kar, and Oleg Rokhlenko.
  2022{\natexlab{a}}.
\newblock {MultiCoNER: a Large-scale Multilingual dataset for Complex Named
  Entity Recognition}.

\bibitem[{Malmasi et~al.(2022{\natexlab{b}})Malmasi, Fang, Fetahu, Kar, and
  Rokhlenko}]{multiconer-report}
Shervin Malmasi, Anjie Fang, Besnik Fetahu, Sudipta Kar, and Oleg Rokhlenko.
  2022{\natexlab{b}}.
\newblock {SemEval-2022 Task 11: Multilingual Complex Named Entity Recognition
  (MultiCoNER)}.
\newblock In \emph{Proceedings of the 16th International Workshop on Semantic
  Evaluation (SemEval-2022)}. Association for Computational Linguistics.

\bibitem[{Meng et~al.(2021)Meng, Fang, Rokhlenko, and Malmasi}]{meng2021gemnet}
Tao Meng, Anjie Fang, Oleg Rokhlenko, and Shervin Malmasi. 2021.
\newblock {GEMNET: Effective gated gazetteer representations for recognizing
  complex entities in low-context input}.
\newblock In \emph{Proceedings of the 2021 Conference of the North American
  Chapter of the Association for Computational Linguistics: Human Language
  Technologies}, pages 1499--1512.

\bibitem[{Miyato et~al.(2017)Miyato, Dai, and
  Goodfellow}]{miyato2017adversarial}
Takeru Miyato, Andrew~M Dai, and Ian Goodfellow. 2017.
\newblock Adversarial training methods for semi-supervised text classification.
\newblock \emph{stat}, 1050:6.

\bibitem[{Morwal et~al.(2012)Morwal, Jahan, and Chopra}]{morwal2012named}
Sudha Morwal, Nusrat Jahan, and Deepti Chopra. 2012.
\newblock Named entity recognition using hidden markov model (hmm).
\newblock \emph{International Journal on Natural Language Computing (IJNLC)
  Vol}, 1.

\bibitem[{Mueller et~al.(2020)Mueller, Andrews, and
  Dredze}]{mueller2020sources}
David Mueller, Nicholas Andrews, and Mark Dredze. 2020.
\newblock Sources of transfer in multilingual named entity recognition.
\newblock In \emph{Proceedings of the 58th Annual Meeting of the Association
  for Computational Linguistics}, pages 8093--8104.

\bibitem[{Sherstinsky(2020)}]{sherstinsky2020fundamentals}
Alex Sherstinsky. 2020.
\newblock Fundamentals of recurrent neural network (rnn) and long short-term
  memory (lstm) network.
\newblock \emph{Physica D: Nonlinear Phenomena}, 404:132306.

\bibitem[{Sui et~al.(2019)Sui, Chen, Liu, Zhao, and Liu}]{sui2019leverage}
Dianbo Sui, Yubo Chen, Kang Liu, Jun Zhao, and Shengping Liu. 2019.
\newblock Leverage lexical knowledge for chinese named entity recognition via
  collaborative graph network.
\newblock In \emph{Proceedings of the 2019 Conference on Empirical Methods in
  Natural Language Processing and the 9th International Joint Conference on
  Natural Language Processing (EMNLP-IJCNLP)}, pages 3830--3840.

\bibitem[{Wu et~al.(2021)Wu, Li, Wang, Meng, Qin, Chen, Zhang, Liu
  et~al.}]{wu2021r}
Lijun Wu, Juntao Li, Yue Wang, Qi~Meng, Tao Qin, Wei Chen, Min Zhang, Tie-Yan
  Liu, et~al. 2021.
\newblock R-drop: regularized dropout for neural networks.
\newblock \emph{Advances in Neural Information Processing Systems}, 34.

\bibitem[{Yadav and Bethard(2018)}]{yadav2018survey}
Vikas Yadav and Steven Bethard. 2018.
\newblock A survey on recent advances in named entity recognition from deep
  learning models.
\newblock In \emph{Proceedings of the 27th International Conference on
  Computational Linguistics}, pages 2145--2158.

\bibitem[{Yan et~al.(2021)Yan, Gui, Dai, Guo, Zhang, and Qiu}]{yan2021unified}
Hang Yan, Tao Gui, Junqi Dai, Qipeng Guo, Zheng Zhang, and Xipeng Qiu. 2021.
\newblock A unified generative framework for various ner subtasks.
\newblock In \emph{Proceedings of the 59th Annual Meeting of the Association
  for Computational Linguistics and the 11th International Joint Conference on
  Natural Language Processing (Volume 1: Long Papers)}, pages 5808--5822.

\bibitem[{Zhang and Yang(2018)}]{zhang2018chinese}
Yue Zhang and Jie Yang. 2018.
\newblock Chinese ner using lattice lstm.
\newblock In \emph{Proceedings of the 56th Annual Meeting of the Association
  for Computational Linguistics (Volume 1: Long Papers)}, pages 1554--1564.

\bibitem[{Zhong and Chen(2021)}]{zhong2021frustratingly}
Zexuan Zhong and Danqi Chen. 2021.
\newblock A frustratingly easy approach for entity and relation extraction.
\newblock In \emph{Proceedings of the 2021 Conference of the North American
  Chapter of the Association for Computational Linguistics: Human Language
  Technologies}, pages 50--61.

\bibitem[{Zhu et~al.(2019)Zhu, Cheng, Gan, Sun, Goldstein, and
  Liu}]{zhu2019freelb}
Chen Zhu, Yu~Cheng, Zhe Gan, Siqi Sun, Tom Goldstein, and Jingjing Liu. 2019.
\newblock Freelb: Enhanced adversarial training for language understanding.

\end{thebibliography}
\bibliographystyle{acl_natbib}




\section*{Appendix}
\appendix
    
  \section{ Hyper parameters setting }
    \label{appendix1}
    In these tables, T, C and L under graph parameter denote T-graph, C-graph and L-graph respectively which means we use one or all of these three graph in our  LERoBERTa-GCN4NER model.
    
    \begin{table}[h]
    \caption{ Scope of each hyper parameters that we have tried in our experiments }
    \centering
    \renewcommand\arraystretch{1}{
    \setlength{\tabcolsep}{1mm}{
    \begin{tabular}{c c}
    \toprule[1.5pt]
    parameters  &   scope     \\
    \hline
    learning rate    &   \{ $ 5 \times 10^{-6}$, $ \{1,3,5 \}\times 10^{-5}$  \}    \\
    warming up rate    &   \{ 0.06, 0.1 \}     \\
    seed    &   \{ 42, 100, 200, 2022 \}    \\
    batch size  &   \{ 8, 16, 32 \}   \\
    epoch  &   \{ 30, 50 \}   \\
    FGM($\epsilon$)    &    \{ 0.1, 0.3, 0.5, 0.8, 1 \}    \\
    PGD($\epsilon$)    &    \{ 0.1, 0.3, 0.5, 0.8, 1 \}    \\
    PGD($\alpha$)   &    \{ 0.3, 1 \}    \\
    PGD($K$)    &    \{ 1, 2, 3, 5 \}    \\
    FreeLB($adv\_lr$)    &    \{ $\{1,3,5\}\times 10^{-5}$ \}    \\
    FreeLB($mag$)    &    \{ 0.05, 0.1, 0.5 \}    \\
    FreeLB($K$)    &    \{ 1, 2, 3, 5 \}    \\
    R-Drop($\alpha$)    &    \{ 0.01, 0.05, 0.2, 0.5 \}    \\
    GCN4NER(graph)    &    \{ T+C+L, T,  C, L \}    \\
    \bottomrule[1pt]
    \end{tabular}}}
    \label{tab.Hyperparameters_setting}
    \end{table}

  \section{Macro-f1 scores(\%) predicted singly by different models under different parameters}
    \label{appendix2}
    
    
    \begin{table*}[h]
    \caption{ Macro-f1 scores(\%) predicted by every single model  under different parameters on Chinese(ZH) test dataset at model ensemble stage }
    \centering
    \renewcommand\arraystretch{1.5}{
    \setlength{\tabcolsep}{4.5mm}{
    \begin{tabular}{c c c c c c c}
    \toprule[1.5pt]
    id & macro-f1 & learning rate & warming up rate  & batch sieze &  PGD  &  graph \\
    \hline
    1    &   \textbf{67.71}    &  $1 \times 10^{-5}$   &  0.06    &   32  &   \textbf{+}    &   T+C+L \\
    2    &   67.56    &  $5 \times 10^{-6}$   &  0.1    &   32  &   \textbf{+}    &   T+C+L \\
    3    &   67.68    &  $1 \times 10^{-5}$   &  0.06    &   16  &   \textbf{+}    &   T+C+L \\
    4    &   67.43    &  $1 \times 10^{-5}$   &  0.06    &   32  &   \textbf{-}    &   T+C+L \\
    5    &   67.35    &  $1 \times 10^{-5}$   &  0.06    &   32  &   \textbf{+}    &   T \\
    6    &   67.32    &  $1 \times 10^{-5}$   &  0.06    &   32  &   \textbf{+}    &   C \\
    7    &   67.21    &  $1 \times 10^{-5}$   &  0.06    &   32  &   \textbf{+}   &   L \\
    \bottomrule[1pt]
    \end{tabular}}}
    \label{tab.zh_singly_ensemble}
    \end{table*}

    
    \begin{table*}[h]
    \caption{ Macro-f1 scores(\%) predicted by every single model  under different parameters on Chinese(ZH) test dataset at pseudo labeling fine-tuning stage }
    \centering
    \renewcommand\arraystretch{1.5}{
    \setlength{\tabcolsep}{4.5mm}{
    \begin{tabular}{c c c c c c c}
    \toprule[1.5pt]
    id & macro-f1 & learning rate & warming up rate  & batch sieze &  PGD  &  graph \\
    \hline
    1    &   72.56    &  $1 \times 10^{-5}$   &  0.06    &   32  &   \textbf{+}    &   T+C+L \\
    2    &   72.89    &  $5 \times 10^{-6}$   &  0.1    &   32  &   \textbf{+}    &   T+C+L \\
    3    &   \textbf{72.93}    &  $1 \times 10^{-5}$   &  0.06    &   16  &   \textbf{+}    &   T+C+L \\
    4    &   72.53    &  $1 \times 10^{-5}$   &  0.06    &   32  &   \textbf{-}    &   T+C+L \\
    5    &   72.58    &  $1 \times 10^{-5}$   &  0.06    &   32  &   \textbf{+}    &   T \\
    6    &   72.42    &  $1 \times 10^{-5}$   &  0.06    &   32  &   \textbf{+}    &   C \\
    7    &   72.33    &  $1 \times 10^{-5}$   &  0.06    &   32  &   \textbf{+}    &   L \\
    \bottomrule[1pt]
    \end{tabular}}}
    \label{tab.zh_pseudo_singly_ensemble}
    \end{table*}

    
    \begin{table*}[h]
    \caption{ Macro-f1 scores(\%) predicted by every single model  under different parameters on multilingual test dataset at model ensemble stage }
    \centering
    \renewcommand\arraystretch{1.5}{
    \setlength{\tabcolsep}{3.6mm}{
    \begin{tabular}{c c c c c c c}
    \toprule[1.5pt]
    id  &   macro-f1    & learning rate & warming up rate  & batch sieze &  PFE-SFE  &  R-Drop($\alpha$) \\
    \hline
    1    &   72.32    &  $5 \times 10^{-6}$   &  0.06    &   16  &   \textbf{+}    &  0.05 \\
    2    &   \textbf{73.18}    &  $5 \times 10^{-6}$   &  0.06   &   16  &   \textbf{+}    &   0.1 \\
    3    &   72.86    &  $5 \times 10^{-6}$   &  0.06    &   16  &   \textbf{+}    &   0.2 \\
    4    &   72.64    &  $1 \times 10^{-5}$   &  0.06    &   16  &   \textbf{+}    &   0.1 \\
    5    &   72.12    &  $5 \times 10^{-6}$   &  0.06    &   16  &   \textbf{-}    &   0.1 \\
    6    &   73.01    &  $5 \times 10^{-6}$   &  0.1    &   16  &   \textbf{+}    &   0.1 \\
    7    &   73.09    &  $5 \times 10^{-6}$   &  0.06    &   32  &   \textbf{+}    &   0.1 \\
    \bottomrule[1pt]
    \end{tabular}}}
    \label{tab.mul_singly_ensemble}
    \end{table*}

    
    \begin{table*}[h]
    \caption{ Macro-f1 scores(\%) predicted by every single model  under different parameters on multilingual test dataset at pseudo labeling fine-tuning stage }
    \centering
    \renewcommand\arraystretch{1.5}{
    \setlength{\tabcolsep}{3.6mm}{
    \begin{tabular}{c c c c c c c}
    \toprule[1.5pt]
    id  &   macro-f1    & learning rate & warming up rate  & batch sieze &  PFE-SFE  &  R-Drop($\alpha$) \\
    \hline
    1    &   76.43    &  $5 \times 10^{-6}$   &  0.06    &   16  &   \textbf{+}    &  0.05 \\
    2    &   \textbf{76.91}    &  $5 \times 10^{-6}$   &  0.06   &   16  &   \textbf{+}    &   0.1 \\
    3    &   76.74    &  $5 \times 10^{-6}$   &  0.06    &   16  &   \textbf{+}    &   0.2 \\
    4    &   76.18    &  $1 \times 10^{-5}$   &  0.06    &   16  &   \textbf{+}    &   0.1 \\
    5    &   75.96    &  $5 \times 10^{-6}$   &  0.06    &   16  &   \textbf{-}    &   0.1 \\
    6    &   76.63    &  $5 \times 10^{-6}$   &  0.1    &   16  &   \textbf{+}    &   0.1 \\
    7    &   76.78    &  $5 \times 10^{-6}$   &  0.06    &   32  &   \textbf{+}    &   0.1 \\
    \bottomrule[1pt]
    \end{tabular}}}
    \label{tab.mul_pseudo_singly_ensemble}
    \end{table*}

    \begin{table*}[h]
    \caption{ Macro-f1 scores(\%) predicted by every single model  under different parameters on mix-code test dataset at model ensemble stage }
    \centering
    \renewcommand\arraystretch{1.5}{
    \setlength{\tabcolsep}{3.5mm}{
    \begin{tabular}{c c c c c c c}
    \toprule[1.5pt]
    id  &   macro-f1    & learning rate & warming up rate & batch sieze &  data augment   &  decoder \\
    \hline
    1    &   \textbf{80.32}    &  $1 \times 10^{-6}$   &  0.06    &   32  &   MR + SiS    &   crf \\
    2    &   80.21    &  $1 \times 10^{-6}$   &  0.06    &   32  &   MR + SiS    &   crf \\
    3    &   79.84    &  $1 \times 10^{-5}$   &  0.06    &   32  &   MR + SiS    &   softmax \\
    4    &   80.09    &  $1 \times 10^{-6}$   &  0.1    &   32  &   MR + SiS    &   softmax \\
    5    &   79.94    &  $1 \times 10^{-6}$   &  0.06   &   32  &   MR    &   crf \\
    6    &   80.11    &  $5 \times 10^{-6}$   &  0.06    &   16  &   MR    &   crf \\
    7    &   79.75    &  $1 \times 10^{-6}$   &  0.06    &   32  &   MR    &  softmax \\
    
    \bottomrule[1pt]
    \end{tabular}}}
    \label{tab.mix_singly_ensemble}
    \end{table*}

    \begin{table*}[h]
    \caption{ Macro-f1 scores(\%) predicted by every single model  under different parameters on mix-code test dataset at pseudo labeling fine-tuning stage }
    \centering
    \renewcommand\arraystretch{1.5}{
    \setlength{\tabcolsep}{3.5mm}{
    \begin{tabular}{c c c c c c c}
    \toprule[1.5pt]
    id  &   macro-f1    & learning rate & warming up rate & batch sieze &  data augment   &  decoder \\
    \hline
    1    &   \textbf{83.91}    &  $1 \times 10^{-6}$   &  0.06    &   32  &   MR + SiS    &   crf \\
    2    &   83.38    &  $1 \times 10^{-6}$   &  0.06    &   32  &   MR + SiS    &   crf \\
    3    &   83.40    &  $1 \times 10^{-5}$   &  0.06    &   32  &   MR + SiS    &   softmax \\
    4    &   83.55    &  $1 \times 10^{-6}$   &  0.1    &   32  &   MR + SiS    &   softmax \\
    5    &   83.00    &  $1 \times 10^{-6}$   &  0.06   &   32  &   MR    &   crf \\
    6    &   83.70    &  $5 \times 10^{-6}$   &  0.06    &   16  &   MR    &   crf \\
    7    &   83.50    &  $1 \times 10^{-6}$   &  0.06    &   32  &   MR    &  softmax \\
    
    \bottomrule[1pt]
    \end{tabular}}}
    \label{tab.mul_pseudo_singly_ensemble}
    \end{table*}

\end{document}